%% file: main.tex
\documentclass[runningheads]{llncs}
\usepackage[T1]{fontenc}
\usepackage{graphicx}
\usepackage{booktabs}
\usepackage[misc]{ifsym}

\usepackage{mwe}
\usepackage{amsfonts}       % blackboard math symbols
\usepackage{nicefrac}       % compact symbols for 1/2, etc.
\usepackage{microtype}      % microtypography
\usepackage{xcolor}         % colors
\usepackage{amsmath} % for math symbols and environments
\usepackage{amssymb}
\usepackage{subfigure}
\usepackage{algorithmicx}
\usepackage{hyperref}  
\usepackage{multirow}

\usepackage{amsfonts}  
\usepackage{url}    
\usepackage{tikz}
\usepackage{pgfplots}
\usepackage{float}
\pgfplotsset{compat=1.17}
\usepackage{algorithm}
\usepackage{algpseudocode}

\algnewcommand\algorithmicinput{\textbf{Input:}}
\algnewcommand\Input{\item[\algorithmicinput]}

\algnewcommand\algorithmicoutput{\textbf{Output:}}
\algnewcommand\Output{\item[\algorithmicoutput]}

\begin{document}

%\title{Uplift Modeling under Limited Supervision}
\title{Uplift Modeling Under Limited Supervision}
%\titlerunning{Uplift Modeling under Limited Supervision}
\titlerunning{Uplift Modeling Under Limited Supervision}

%\author{Anonymous Authors}
%\institute{~}
% If the full title of your paper is short enough to also fit in the running head, you can omit the abbreviated paper title here. You can check as follows: if you comment out the \titlerunning line, something will appear in the header of all odd-numbered pages of your PDF from page 3 onward. This something is either the full title (in which case all is well), or the error message "Title Suppressed Due to Excessive Length". If this error message appears, you're going to want to provide an abbreviated title within the \titlerunning command, because if you won't do it, Springer will do it for you.

%N.B.: Author information (both in the \author{} and \authorrunning{} command) should only be present in the Camera-Ready Version of your paper. The version that you initially submit for review, ought to be double-blind. So, when initially submitting your paper, use:
%\author{Author information scrubbed for double-blind reviewing}
\author{George Panagopoulos{\Letter}\inst{1}%\orcidID{0000-0001-7731-9448} 
\and
Daniele Malitesta \inst{2} %\orcidID{0000-0003-2228-0333} 
\and
Fragkiskos D. Malliaros \inst{2} %\orcidID{0000-0002-8770-3969}
 \and
Jun Pang\inst{1} %\orcidID{0000-0002-4521-4112} 
}
%Antwan Andr\'e Patton\inst{1}\orcidID{0000-1111-2222-3333}}
% You may leave out the orcidID information, if you want to.
% Use \corr to indicate the corresponding author. Note the spacing around the \corr command. Only one author can be the corresponding author.

%N.B.: comment out the \authorrunning{} command for the double-blind version of your paper submitted for review. Later, if your paper is accepted, use the command for the Camera-Ready Version.
\authorrunning{G. Panagopoulos et al.}
% If there are two authors, write 'A.L. Benjamin and C.C. Broadus Jr.'
% If there are more than two authors, '[...] et al.' is used.

\institute{University of Luxembourg, 6 avenue de la Fonte, Esch-sur-Alzette, Luxembourg \\ \texttt{\{georgios.panagopoulos, jun.pang\}@uni.lu} 
\and 
Université Paris-Saclay, CentraleSupélec, Inria, 3 Rue Joliot Curie, Gif-sur-Yvette, France\\ \texttt{\{daniele.malitesta, fragkiskos.malliaros\}@centralesupelec.fr}}

%\and
%Fictional West Coast University, Long Beach CA 90840, USA \email{ccb@fwcu.fake}
%\and
%Secondary European Affiliation, Tiergartenstr. 17, 69121 Heidelberg, Germany
%\email{lncs@springer.com}}

\maketitle              % typeset the header of the contribution
\begin{abstract}
%Traditional causal inference relies on randomized control trials that require interventions to estimate the treatment effect. When interventions are costly, machine learning can be used to predict the treatment effect on the population using a trial on a subset of samples for supervision, commonly referred to as uplift modeling.
 %Although the core motivation is to reduce the risk of interventions, current uplift modeling methods rely on a significant subset of the data to undergo such a trial in order to form a training set. 
%To limit the need for supervision, we propose an active semi-supervised uplift modeling method where we utilize the underlying connections of the samples with bipartite graph neural networks. Additionally, we develop an acquisition function based on uncertainty quantification and sample diversity to build the training set under a budget constraint, in an active learning fashion. 
%We evaluate our model on large-scale real-world networks with both ground truth and semi-simulated experimental variables. 
%Our experiments indicate that the proposed methodology consistently outperforms the existing methods from uplift modeling and individual treatment effect literature, highlighting the potential of this framework to diminish the risk of excessive testing. 
Estimating causal effects in e-commerce tends to involve costly treatment assignments which can be impractical in large-scale settings. Leveraging machine learning to predict such treatment effects without actual intervention is a standard practice to diminish the risk. However, existing methods for treatment effect prediction tend to rely on training sets of substantial size, which are built from real experiments and are thus inherently risky to create. In this work we propose a graph neural network to diminish the required training set size, relying on graphs that are common in e-commerce data. Specifically, we view the problem as node regression with a restricted number of labeled instances, develop a two-model neural architecture akin to previous causal effect estimators, and test varying message-passing layers for encoding.
Furthermore, as an extra step, we combine the model with an acquisition function to guide the creation of the training set in settings with extremely low experimental budget. The framework is flexible since each step can be used separately with other models or treatment policies. The experiments on real large-scale networks indicate a clear advantage of our methodology over the state of the art, which in many cases performs close to random, underlining the need for models that can generalize with limited supervision to reduce experimental risks.

%that the network improves the model’s predictive ability, while an active learning policy significantly enhances the final ranking.

%Traditional causal inference relies on randomized control trials that require interventions on half of the sample set to derive an estimate of the treatment's effect.  When interventions are costly, time-consuming, or even impossible, machine learning can be used to predict the effect of the treatment. A limited experiment is performed and the outcome is used to train a model to predict the average treatment effect in the rest of the population. The methodology is called uplift modeling and has a broad range of applications, from healthcare and social science to e-commerce and marketing. Although uplift modeling's core motivation is to diminish the number of interventions, the current methods rely on models that require a significant subset of the data to undergo such a trial to form a training set. In this work, we develop a semi-supervised learning method that diminishes the number of required tests. We rely on the underlying connections of the samples to uncover confounders that enhance the ability of the model to generalize. Our preliminary experiments indicate that the network contains information that is crucial to enhance the model's prediction.

\end{abstract}
\input{introduction}
\input{related}

\input{methodology}

\input{experiments}

\input{conclusion}

\bibliographystyle{splncs04}
\bibliography{upgnn}
\input{appendix}
%\input{appendix}
%% Note that this preceding line implies that you store your BibTeX references in a file called 'mybibliography.bib'. If you instead store your references in a file with a different name, for instance 'references.bib', the preceding line should read '\bibliography{references}'. Whatever you do, DO NOT put the file name extension .bib inside the \bibliography command; this will trip up LaTeX compilers. 
%
% If you do not want to use BibTeX, you can also type up the bibliography exactly as you see fit, using the following structure:

\end{document}

%% file: introduction.tex
\section{Introduction}
\label{sec:introduction}
Statistical tests for causal inference are ubiquitous, from drug discovery~\cite{tye2004application} and psychology~\cite{wright2006comparing} to social studies and online platforms---being at the core of decision-making. 
A prevalent example is A/B testing~\cite{gui2015network}, the de facto way to evaluate the potential of a system change before applying it in scenarios such as e-commerce.
Typically, the population is split into two random groups, and the treatment is assigned to one of them ($T=1$). The difference between their response variable $Y$ (e.g., time spent on the system) quantifies the average treatment effect (ATE) of the randomized control trial (RCT)~\cite{rubin1974estimating}, i.e., $Y_T - Y_C = \mathbb{E}[Y|T=1] - \mathbb{E}[Y|T=0]$, where $T$ and $C$ refer to the treatment and control groups, respectively.  
In this setting, however, we risk causing churn on a massive scale if the proposed change is not effective \cite{bakshy2012social}. It is thus more wise to test on a smaller scale and try to extrapolate our findings. 
%Alternatively, consider an experiment that studies the effects of smoking on a disease. Since it is unethical to force a subject to start or quit smoking, the treatment assignment is limited accordingly.
% marketing campaign where the company gives away promotion coupons to attract more attention. A coupon is the treatment and the outcome is the increase  but will go bankrupt if every user gets a coupon.
In these cases, we can build a model that predicts the outcome without actually intervening, based on the samples' confounders, e.g., for a user, this could be the %frequency and duration of visits, 
purchase history and demographics. 
The core hypothesis is that given the common assumption of unconfoundedness~\cite{dawid1979conditional}, samples with similar confounders will exhibit similar outcomes under the same treatment. 
%, where we assume that the outcome happens only because of the treatment, when we have observed all the possible confounders between treatment and outcome.
Developing a model that can predict the effect of a treatment on unseen samples is referred to as \textit{uplift modeling} (UM)~\cite{radcliffe2011real,betlei2021uplift}, mainly stemming from business context applications. The uplift refers to the prediction of the outcome's change due to the treatment, i.e., the ATE, for a set of samples~\cite{diemert2018large}. 
The problem is akin to individual treatment effect (ITE) estimation, i.e., predicting the  outcome of the same sample with and without treatment (counterfactual) and computing their difference. %$E[Y_i|X_i, T_i = 1] - E[Y_i|X_i, T_i = 0]$ based on the covariates. 
%\textcolor{red}{ Note that this work is different from the existing studies on spillover effect, also known as interference, where the observed treatment or outcome of an instance may causally influence the outcomes of the connected instances. In contrast, we focus on the situations where the network structure can be exploited for controlling confounding bias. }

In this work, our attention is directed towards a real-world scenario of a marketing campaign where promoting is costly/risky/time-consuming---a typical uplift modeling setting \cite{diemert2018large,olaya2021or}. In this context, we aim to rank the whole user base based on how they will respond to the campaign, i.e., how much more they will consume if they receive a coupon. The coupon resembles a treatment, and the difference in consumption is the effect.  %i.e. have opposite outcomes for a binary $Y$ or the biggest difference between $T=1$ and $T=0$ for a continuous. 
%This obviously differs significantly from simply ranking users in terms of predicted outcomes without taking treatment into account, because the outcome may depend on several exogenous factors and does not reveal the causal relationship. 
We assume there is a fixed budget for the number of users that can participate in the experiment and a predefined random balanced treatment allocation to diminish the risks and costs. 
%Hence the ranking is evaluated based on the ground truth difference mentioned above, and not in $E[Y]$.
The problem can then be broken down into two subproblems:
\begin{itemize}
\item Which users should participate in the experiment?
\item How to use this experiment to predict the outcome of the rest of the dataset?
\end{itemize}
We approach the first problem with an active learning formulation~\cite{settles2009active}, where we sequentially choose subsets of the samples to label in order to maximize the model's effectiveness until we reach our budget.
The latter problem pertains to the model's capacity to generalize with limited supervision, which can be cast as a semi-supervised learning problem where graph neural networks (GNNs) are rather effective~\cite{kipf2016semi}. 
% utilize the relational structure of samples to facilitate generalization when the training labels are scarce. 
In general, GNNs are popular in social network applications and can be combined effectively with decision algorithms \cite{panagopoulos2023maximizing}. 
The majority of online systems can be represented as heterogeneous graphs, e.g., user-product purchases (e-commerce~\cite{DBLP:conf/sigir/Wang0WFC19}) %DBLP:conf/sigir/0001DWLZ020
or follow relationships (social media~\cite{DBLP:conf/www/Fan0LHZTY19}).%,DBLP:conf/www/QuanDGYJL23
Thus, we frame the problem as an active semi-supervised learning task on a bipartite graph and address it in alternating rounds. %\smallskip\noindent{\bf Contributions.}

\smallskip\noindent{\bf Contributions.} Our contributions can be summarized as follows:
\begin{enumerate}
    \item We develop a novel modular framework, based on two steps, a GNN, and an active learning method, to address the need for limited supervision in UM, moving from the standard ``70\%-80\%'' train set rule~\cite{devriendt2018literature,gutierrez2017causal} down to 5\%-20\%.
    \item We formulate UM for networks and test it in an open, large-scale network with real experimental annotations. To the best of our knowledge, this is the first such attempt in the literature. Moreover, we focus on continuous outcomes, which are relatively understudied compared to binary UM~\cite{verhelst2023uplift} but equally prevalent in the real world.
    \item We conduct experiments with models from the UM and ITE literature, including neural, tree-based, and graph-aware methods, and showcase that the proposed methodology surpasses the state-of-the-art substantially.
\end{enumerate}
%The training set size i.e. the budget is central in our analysis, as is in the real world. The vast majority of UM literature assumes no specific constraint in supervision and utilizes the common . 

 %In contrast, we evaluate based on uplift ranking~\cite{rafla2022evaluation} which are prevalent in commercial applications.
    %\item Our target is to utilize semi-supervised and active learning as a means to diminish the supervision, rather than debiasing the observed outcome due to interference. Hence our approach runs in tandem with the experiment and as the train set increases not after it.
    %\item  Submodular maximization is utilized extensively in optimal experimental design\cite{golovin2010near,wei2015submodularity,jagalur2021batch,tigas2022interventions} to derive intervention assignment policies. However, their target is to improve the study of a given hypothesis. Our active learning policy aims to find samples that will improve UM and subsequently user ranking.
%Our initial results indicate that a heterogeneous graph neural network outperforms the state-of-the-art in several uplift modeling metrics.
%\textcolor{red}{Diemert: Spend is convex wrtto Bid (with some jumps – increased bids make you win more auctions) Uplift is concave wrtto Bid (diminishing returns) So we need to devise a much more complex system than just ranking individuals by uplift !}

%% file: related.tex
\section{Related Work}
\label{sec:related}
%\subsection{Machine Learning in Causality}
One of the most prevalent applications of machine learning (ML) in causality is estimating the probability of a sample being assigned to a treatment, i.e., the propensity scores using the confounders, which can uncover design bias~\cite{lee2010improving}. 
Models that predict the propensity score and the outcome can be used in tandem in the double ML framework~\cite{chernozhukov2017double}, which is provably doubly robust (DR), in the sense that it is consistent if either the propensity or the outcome predictive model is consistent.  %These results indicate that the epistemic uncertainty introduced by the models can be diminished~\cite{chernozhukov2018double}.
DR has been utilized for heterogeneous causal effect estimation \cite{kennedy2023towards}, which is similar to UM.
Simpler models use treatment as an extra input (\textsc{S-Learner}). Two models that learn the output of treatment and control groups (\textsc{T-Learner}) are prevalent~\cite{radcliffe2007using} and are the basis for the successful causal meta-learners (R/X-Learner)~\cite{kunzel2019metalearners}. 
In addition, meta-learners have been utilized in the context of cost-effective uplift modeling \cite{vanderschueren2024metalearners}.
%, which have also exhibited impressive results in trials with no prior related data~\cite{nilforoshan2023zero}. 

Class transformation with regularized Logistic Regression~\cite{rudas2023regularization} or \textsc{XGBoost}~\cite{soltys2018boosting} is also effective for binary outcomes and under balanced treatment assignments, while random forests were some of the first models developed for causal effect estimation \cite{wager2018estimation}. Causal estimator models that allow for instrumental variables have been developed based on deep neural networks \cite{hartford2017deep}. % treatment is imbalanced, reweighing strategies have to be employed~\cite{athey2015machine}.  
\textsc{UpliftTrees}~\cite{rzepakowski2012decision} have custom splitting criteria based on the outcome distribution in the treatment groups of the tree nodes.

The effect of the network has not been examined in the context of UM, but it has for ITE. 
One of the first works that proposed an adjustment to the causal effect estimation based on the network was Arbour et al.~\cite{arbour2016inferring}.
Veitch et al.~\cite{veitch2019using} developed a neural method to identify hidden confounders in interconnected samples. They extend the doubly robustness for non-iid data and build a model that relies on random-walk-based node representations. 
The same team proposed \textsc{Dragonnet}~\cite{shi2019adapting}, a neural architecture that is split into predicting the outcomes under each treatment and the propensity score. %neural models that learn common representations for both types of outcomes, such as 

Guo et al.~\cite{guo2020learning} developed an ITE model with GNN encoding in the initial layers and split output layers for each treatment. %The model receives the network, confounders, and the treatment as input and outputs two scores from different layers representing the outcome for each treatment. 
Furthermore, it minimizes the distance between the representations' distribution under different treatments. This stems from a seminal work on bounding the generalization error of ITE models by the sum of the given model's standard error and the distance between the treatment and control confounders' distributions~\cite{shalit2017estimating}.   
Similar GNNs have been developed based on multi-task and adversarial learning~\cite{chu2021graph,jiang2022estimating}, the geometric curvature of the network~\cite{farzam2023curvature}, or a network with multiple relations~\cite{lin2023estimating}. 
It should be noted that predicting the causal estimates under network confounding and the network interference \cite{ma2021causal,cristali2022using} differs conceptually, although the models are similar.
An effort to predict ITE considering both the network confounding and the interference was made using hypergraph neural networks~\cite{ma2022learning}. 
We refer the reader to Appendix A, where we make a distinction between confounding and interference and justify why we follow the literature~\cite{veitch2019using,chu2021graph,farzam2023curvature}%ma2021deconfounding
on the assumption for the latter.

One important difference between UM and ITE is that the latter defines distinctly the counterfactual prediction and requires the respective ground truth for evaluation, which renders the use of simulated data imperative~\cite{ma2021causal}. Hence, the aforementioned works on network-based ITE are evaluated with simulated experiments on observational data and hardcoded network confounding. Our work uses, for the first time, an open network with ground-truth experimental variables. Furthermore, we address a network with heterogeneous nodes, a setting prevalent in e-commerce. HINITE~\cite{lin2023estimating}, which uses one type of node but multiple relations, is in a similar heterogeneous direction but not comparable.

%% file: methodology.tex
\section{Proposed Methodology}
\label{method} 

We consider the e-commerce scenario where data is commonly structured through bipartite, undirected graphs e.g. user-product. Given the effectiveness of GNNs in semi-supervised learning~\cite{hamilton2017inductive}, we create a general framework that can be tested with several GNN layers. The model can be used as is, but we additionally define an active learning method to build the training set iteratively based on the model's uncertainty and the samples' properties.
 %where the model output serves as input to the acquisition function that builds the training set iteratively. The two steps are modular in the sense that they can be used separately and hence the first is also evaluated as a standalone model in hte experiments.
%The two steps are modular in the sense that they can be used separately and hence the first is also evaluated as a standalone model in hte experiments.
In the following, the scalars are represented by a lowercase letter, the vectors or sets with an upper case, and the matrices with an upper case bold letter.

%Our methodology follows standard notation where the scalars are represented by a lower case letter, the vectors or sets with an upper case and the matrices with an upper case bold letter.

\subsection{Uplift Modeling with Graph Neural Networks (\textsc{UMGNet})}%Two-model Bipartite Graph Neural Network (TBGNN)}

Let $u \in U$ and $p \in P$ be a user and a product, respectively, having $|U| = n$ users and $|P| = m$ products in total. Then, let $\mathbf{R} \in \mathbb{R}^{n \times m}$ be the user-product interaction matrix, where $R_{up} = 1$ if there is a recorded interaction between user $u$ and product $p$, 0 otherwise. Moreover, let each user $u$ be represented by a tuple $Z_u = (\mathbf{X}_u, T_u, Y_u)$ where $\mathbf{X}_U \in \mathbb{R}^{n \times d}$  are the covariates representing features of dimensionality $d$ from all $n$ users, while $Y_U \in \mathbb{R}^n$ are the continuous trial outcomes, and $ T_U \in \{0,1\}^n$ is the treatment assignment.

Based on the Neyman-Rubin framework of potential outcomes~\cite{rubin2005causal} and using the \textit{do} operator introduced by Pearl~\cite{pearl2009causality}, we define the average treatment effect for the population as:

\begin{equation}
    \text{\textit{ATE}} = \mathbb{E}[Y_U| \text{\textit{do}}(T_U=1)] - \mathbb{E}[Y_U| \text{\textit{do}}(T_U=0)].
\end{equation}

\begin{figure}[h]
    \centering
    \includegraphics[scale=0.35]{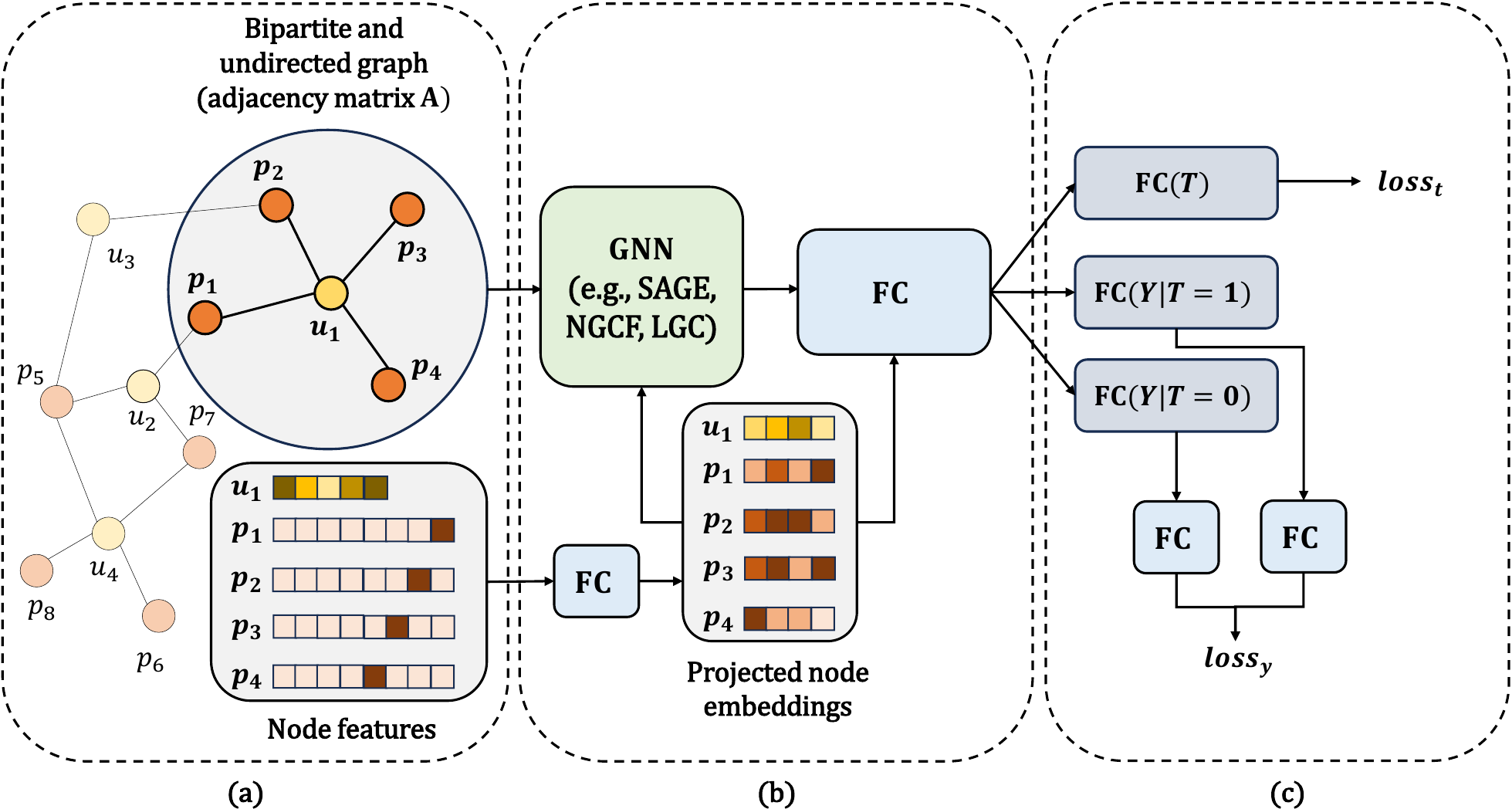}
    \caption{Schematic representation of \textsc{UMGNet}. First \textbf{(a)}, the bipartite and undirected user-product graph, along with the node features, are injected into the framework. Second \textbf{(b)}, node features for users and products are projected to the same latent space through an FC layer and used as input to the GNN model; another FC layer takes the GNN's output and input to predict the regression outcome. Third \textbf{(c)}, outputs for $T = 1$ and $T = 0$ are considered separately and injected into two different FC layers to calculate $\text{\textit{loss}}_y$; the general output of the regression FC is used to calculate $\text{\textit{loss}}_t$.}
    \label{fig:UMGNet}
\end{figure}

\noindent If we consider the law of total expectation and the unconfoundedness~\cite{dawid1979conditional}, i.e., that the confounders $\mathbf{X}$ block all possible ways that the outcome depends on treatment assignment, we can contend that the observed change is an unbiased causal effect from the treatment.
% knowledge of the treatment assignment T does not provide any additional information about the outcome Y
Taking into consideration once again 
%the graph $\mathbf{A}$ of the samples 
the user-product interaction matrix $\mathbf{R}$ as an extra confounder, we have:
\begin{equation}
    \text{\textit{ATE}} = \mathbb{E}[Y_U| \mathbf{X}_U,\mathbf{R}, T_U=1] -  \mathbb{E}[Y_U| \mathbf{X}_U, \mathbf{R}, T_U=0].
\end{equation}

Training one model to predict $\mathbb{E}[Y_U| \mathbf{X}_U, \mathbf{R}, T_U]$, e.g., the \textsc{S-Learner}~\cite{kunzel2019metalearners}, has a specific disadvantage: we expect that $Y$ and $\mathbf{X}$ differ between $T=0$ and $T=1$. %, which would render the model training harder. 
To this end, the two-model methods have exhibited improved performance~\cite{kunzel2019metalearners}, especially in scenarios with imbalanced assignments.
However, the two model approaches do not facilitate sharing the obvious knowledge overlaps. As mentioned in Section~\ref{sec:related}, this can be alleviated by developing a common neural architecture for the first layers~\cite{johansson2016learning}, including treatment prediction~\cite{shi2019adapting} and random walk-based confounders~\cite{veitch2019using}. % that uses separate output layers for each type of treatment that the sample has undergone~\cite{johansson2016learning}. Including treatment prediction and random walk-based confounders improves the results in ITE prediction~\cite{shi2019adapting,veitch2019using}.
%For the same-subject counterfactual predictions, let $Y(1)$ indicate the output under treatment and $Y(0)$ the opposite, we have:
%\begin{equation}
%    Y_i = T_i Y_i(1) + (1-T_i)Y_i(0) 
%\end{equation}
That motivates us to adapt \textsc{Dragonnet}~\cite{shi2019adapting} for bipartite graphs with GNN encoding.

We define the user-product bipartite and undirected graph as %$G=(V,P,E)$ where $V$ are users of size $n$, $p$ are the products of size $m$ and $E$ are the edges. 
$G = (U, P, \mathbf{A})$, where we already introduced $U$ and $P$, while $\mathbf{A} \in \mathbb{R}^{(n + m) \times (n + m)}$ is the adjacency matrix of $G$: 
%Given $\textbf{A}$ the bipartite graph representing the interaction matrix~\cite{chen2022review}, we obtain the adjacency matrix similar to~\cite{li2020hierarchical}: 
\begin{equation}
\label{eq:adj}
    \textbf{A} =  \begin{pmatrix} 0 & \textbf{R}\\ \textbf{R}^\top &0  \end{pmatrix}. 
\end{equation}
% where $\hat{\textbf{A}}  \in \mathbb{Z}^{N+M \times N+M}$.
%Apart from the standard users features, we extract also the graph-related features from the \underline{transaction before the treatment} such as the average purchase price, number of products, and average count. In addition, we perform label prop the train set labels through before treatment AND train set users graph.

\noindent The users' features are represented as $\mathbf{X}_U \in \mathbb{R}^{n \times d}$, and the product features as $\mathbf{X}_P \in \mathbb{R}^{m \times m}$. %as their representation is not strictly fundamental to the rationale of our framework. 
First, we project users and products to the same latent space through a fully connected (FC) layer. The obtained projections are horizontally concatenated to get a common set of node embeddings $\mathbf{X} \in \mathbb{R}^{(n + m) \times w}$, as follows: 
% The neural architecture starts with a feed-forward layer that embeds the user features mentioned above and the product features which are one-hot vectors. 
% The embedding layers for both have the same dimensionality, thus we can concatenate their outputs them \underline{horizontally} into one common $\hat{\textbf{X}}$.
% \begin{equation}
% \label{eq:concat}
%     \hat{\textbf{X}} = [ReLU(\mathbf{X_u}\mathbf{W_u}),ReLU(\mathbf{X_p} \mathbf{W_p})]
% \end{equation}
\begin{equation}
\label{eq:concat}
    \mathbf{X} = \text{concat}\big([\text{ReLU}(\mathbf{X}_U \mathbf{W}_U), \text{ReLU}(\mathbf{X}_P \mathbf{W}_P)]\big),
\end{equation}
where $\mathbf{W}_U \in \mathbb{R}^{d \times w}$ and $\mathbf{W}_P \in \mathbb{R}^{m \times w}$.

Subsequently, we learn graph features leveraging  GNN layers. For this part, we have examined different architectures, including GraphSAGE~\cite{hamilton2017inductive}, NGCF~\cite{DBLP:conf/sigir/Wang0WFC19}, and LGC~\cite{DBLP:conf/sigir/0001DWLZ020}, so we are going to keep it general in terms of formulation:
% \begin{equation}
%      \mathbf{H}_{1} = ReLU( [ \hat{\mathbf{X}}, \hat{\mathbf{A}} \, \hat{\mathbf{X}} \, \mathbf{W}_0]) 
%  \end{equation}
 \begin{equation}
     \mathbf{H}_{1} = \text{GNN}(\mathbf{A}, \mathbf{X}). 
 \end{equation}

\noindent The representations are then broken into two separate paths for $T=1$ and $T=0$ denoted as $t$ and $c$, respectively, through another $\text{FC}$ layer. We utilize a residual connection from $\mathbf{X}$, arriving for $T=1$ in:
\begin{equation}
    %\mathbf{H}^{i+1} = ReLU([\mathbf{H}, \hat{\mathbf{A}} \, \mathbf{H}^i \, \mathbf{W}^{i+1})
    \mathbf{H}^{t}_{i+1} = \text{ReLU}\big(\big[\mathbf{X}[0:n] ,\mathbf{H}_1[0:n] \big] \mathbf{W}^t_i \big),
\end{equation}
where we are slicing the matrices to keep the first $n$ rows that correspond to the user representations, assuming the horizontal concatenation in Eq.~\eqref{eq:concat}.
% Given the desired depth $F$ the model always outputs two predictions, 
Two $\text{FC}$ layers (with $F$ as depth) predict the outcome under treatment $t$ and control $c$:
\begin{align*}
\hat{Y}^c = \mathbf{H}^{c}_{F} \mathbf{W}^c_{F}, \quad\quad& 
\hat{Y}^t = \mathbf{H}^{t}_{F} \mathbf{W}^t_{F}.
\end{align*}
The loss takes into consideration only the factual outcomes, i.e., where the treatment vector $\mathbf{T}$ has $1$ for the $t$ output and $0$ for the $c$: 
\begin{equation}
\label{eq:y_loss}
\text{\textit{loss}}_y =  \frac{1}{n} \sum_{i=0}^{n}\big( T_i(\hat{Y}^t_i - Y_i)^2 + (1-T_i) ( \hat{Y}^c_i - Y_i)^2\big).
\end{equation}

\noindent We refer to this generic GNN-based uplift modeling framework as \textsc{UMGNet}. A schematic representation of the model is shown in Fig. \ref{fig:UMGNet}. As mentioned above, we have examined various instances of the model with different GNN architectures. 
Lastly, inspired by \cite{shi2019adapting}, we consider a variant denoted by \textsc{UMGNet-Dr}, in which we add an extra output layer that predicts the treatment. For this model, we add the following loss term to the one in Eq.~\eqref{eq:y_loss}:
\begin{equation}
    \text{\textit{loss}}_t = \text{CrossEntropy}\big(T, \text{Sigmoid}(\mathbf{H}^\top_F \mathbf{W}^\top_F)\big).
\end{equation}

\subsection{Active Learning for Uplift GNNs (\textsc{UMGNet-AL})}
% improve screening classifiers \cite{wang2022improving}.
% gnn bayesian optimization for molecular spaces \cite{wang2023graph}
% https://lilianweng.github.io/posts/2022-02-20-active-learning/

Active learning relies on the structure of the data and the uncertainty of the model to build iteratively the train set in scenarios where data labeling is costly~\cite{settles2009active}. 
We can break it down into two parts: the first is the uncertainty estimation and the acquisition function over non-labeled samples, while the second is the active learning policy we use to gather new samples to label, i.e., to test in our case. 

\subsubsection{Uncertainty estimation.}
Uncertainty estimation in graph learning is an active research topic~\cite{stadler2021graph,huang2023conformalized_gnn}.
%Particularly in active learning for molecular tasks, GNNs have been combined with evidential, ensemble regression, gaussian processes, or Monte Carlo dropout for UE~\cite{graff2021accelerating,soleimany2021evidential,wollschlager2023uncertainty}
%Moreover, recently conformal prediction has been extended for graph tasks~\cite{huang2023conformalized_gnn}, but with limited applications as of yet.
The uncertainty can be distinguished based on its source: the model (epistemic) or the data (aleatoric). %Epistemic uncertainty is notably challenging as the samples are related and hence, theoretically interfere~\cite{stadler2021graph}.
We will employ an unprincipled yet effective practice to quantify epistemic uncertainty by measuring the variance on the responses of an ensemble of models or simply performing dropout multiple times~\cite{graff2021accelerating}. The dropout mask is random during inference, so the model will produce different outputs for each test sample, arriving at a Bayesian approximation~\cite{gal2016dropout} of the uncertainty.% and is akin to query by committee in active learning.

%Other UE methods include evidential regression, i.e., learning to regress the parameters of the distribution~\cite{soleimany2021evidential,hirschfeld2020uncertainty} or the GNN's gradient~\cite{wang2023graph}. The latter is not developed for node-level tasks while the former is left for future work.
%Moreover, conformal prediction allows for finite-sample guarantees and has been adjusted recently to non-exchangeable samples such as graphs~\cite{huang2023conformalized_gnn}; however, the lack of data in our setting elicits the construction of a calibration set prohibitive. 

%Maybe an LP $C=-\sum_i a*s_i+ (1-a)*f_i$, $s.t. \sum_i s_i< B , \sum_i kernel( y_i,\sum_j x_j*y) $. $B$ is the training size and $y$ are the features and $kernel$ a similarity function.}

Aleatoric uncertainty can be measured using the structure of the graph and the feature distance. Acquisition functions based on diverse criteria, including both types of uncertainties, have proven more effective in node-level tasks~\cite{wu2019active,gilhuber2023diffusal}. We adopt a similar approach and define $D_u, \forall u \in U $ to be the degree, %i.e., the number of purchases before treatment representing the importance of the user,
$Q_u$ is the model's uncertainty, and $b$ is the budget for new training samples in each iteration. According to the literature~\cite{wu2019active,gilhuber2023diffusal}, we define diversity based on feature clustering. %, \textcolor{blue}{but we did not use propagated features because they have not been examined in bipartite networks and because the available graph before treatment is significantly smaller than the real graph. }
Specifically, we cluster the samples with a $k$-means algorithm in a predefined number of clusters. $C_u$ is the set of samples included in the cluster of sample $u$. It is used to balance the number of samples from each cluster in each batch. We calculate the distance $M_u \in \mathbb{R}^n$ between each sample and its cluster centroid and compute the budget for each cluster based on its relative size and $b$, in $\frac{C_u}{n}b \in \mathbb{R}^{n} $. 
We can cast the problem of choosing a batch to add to the training set as ranking based on $U$, $D$, and $M$, with the first constraint being on the batch size. The second constraint, which represents diversity, pertains to how many samples of a given cluster should be included in the train set in each round. Finally, we want to keep the train set balanced in terms of treatment and control samples, hence the third constraint:
%.
\begin{align*}
    %x_i &= \begin{cases} 1, & \text{if node}\quad i \in V \text{ is chosen} \\
    %                     0, & \text{otherwise}
    %       \end{cases} \\
    \text{maximize} \quad & O = \sum_{u=1}^{N} x_u(Q_u + D_u + M_u)\\
    \text{subject to}:&\sum_{ u = 1}^n x_u \leq b \\
    & \sum_{ j \in C_u} x_j \leq \frac{|C_u|}{n}b \\
    & \sum_{ u = 1}^n x_u T_u \leq \frac{b }{2}\\
    &x_u \in \{0, 1\}, \quad \text{for } u = 1, 2, \ldots, n.
\end{align*}

\noindent The problem can be solved greedily by ranking and feasibility check, as the cluster constraints to not interact between clusters. We solve it in each iteration of batch selection, while $D$, $C$, and $C_b$ are precomputed. In practice, we weigh each of the three criteria in the objective function with a coefficient in $[0,1]$ based on the results of validation.

\subsubsection{Acquisition policy.} Thompson sampling \cite{russo2018tutorial}, and upper-confidence bandits~\cite{garivier2011upper} are some of the most popular policies to perform active learning~\cite{settles2009active}. The lack of knowledge at the initial iterations justifies the use of stochastic policies.
However, greedy selection has been effective in batch active learning, and it is actually provably near-optimal in some cases~\cite{wei2015submodularity}. Since our acquisition function does not rely solely on the model's output, which is less trustworthy due to the initial lack of samples, we are going to use a greedy policy and solve the ranking problem in every iteration. The whole procedure can be seen in Algorithm \ref{alg:cap} for $5$ iterations, assuming \textsc{UMGNet} includes by default the parameters $\mathbf{X}, T, Y, \mathbf{A}$ for clarity.

\begin{algorithm}[t]
\label{alg:al}
\caption{\textsc{UMGNet AL}}\label{alg:cap}
\begin{algorithmic}
\Input $O, D, M, b, \text{\textsc{UMGNet}}$
\State $S \gets \{\text{argmax}_{s\subset U} O(D,M,T,b)\}$
\While{$|S| \leq  5*b$}
%\If{$N$ is even}
    \State $\text{Train} ~ \text{\textsc{UMGNet}}(S)$
    \State $Q, \hat{Y} \gets \text{\textsc{UMGNet}}(U\mathbin{\backslash}S)$
    \State $S \gets S \cup \{\text{argmax}_{s\subset U} O(Q,D,M,T,b)\}$
\EndWhile
\Output $\hat{Y}$
\end{algorithmic}
%\State Return \hat{Y}
\end{algorithm}

% ,tigas2022interventions}.
%In our case, we develop a greedy method and an e-greedy that uses random batches with probability $\epsilon$.

%\begin{algorithmic}[2]
%    \Procedure{\textsc{AL}}{$D,\lambda,\ell$}
%\State set $q \gets [~],\mathcal{S} \gets \emptyset $
%\EndProcedure
%\end{algorithmic}
%\end{algorithm}

%% file: experiments.tex
\section{Experimental Evaluation}
\label{sec:experiments}

We report on the experimental results to empirically justify the soundness of our framework. First, we present the main datasets and how we built them. Second, we outline the benchmarking models adopted for this work. Finally, we discuss the obtained results and answer different questions with an ablation study. The code to reproduce the analysis is in GitHub \footnote{\url{https://github.com/geopanag/UMGNet}}.

\subsection{Datasets}
\label{subsec:datasets}
\subsubsection{RetailHero.} The \href{https://ods.ai/competitions/x5-retailhero-uplift-modeling/data}{\textsl{RetailHero}}~\cite{rafla2022evaluation} dataset is comprised of two equal groups of anonymized users undergoing a marketing campaign, along with their product purchase history and some relevant features. The treatment group has received promotional SMS texts and the binary outcome corresponds to whether the user has made a purchase or not after the promotional SMS. %Since we do not have the outcome of the test set, we will use only the train set.  
We follow suit from the literature and the original competition and utilize features such as age, sex, coupon issue time, coupon redeem time, and delay between issue and redeem time. The products are represented by one-hot encoding. We filter the data for erroneous samples e.g., when delay time is negative, or users are less than 16 years old.
We created a user-product graph based on the purchases performed before the time that the promotional coupon was redeemed from treated users. Note that issue and redeem timestamps are added by the organizers for untreated users as well, and they follow similar properties with the redeem time for treated users. %The distribution between the first purchase and the redeem time is almost identical between the treated and non-treated users e.g. the means are 166 and 21 hours and 166 and 8 hours.
Similar to a real-life experiment, purchases done after the intervention are not accessible in the initial form of the graph, except for training samples, i.e., users that have indeed taken part in the experiment in real life. We set two types of continuous outcomes $Y$: 
 \begin{itemize}
     \item \textsl{RHC}: The difference between the average money spent before and after coupon redeem time. $\text{\textit{ATE}} = 2.60, \bar{Y} = 266, \sigma(Y) = 521$.
     \item \textsl{RHP}: The average money spent after coupon redeem time. $\text{\textit{ATE}} = 1.95, \bar{Y} = 423, \sigma(Y) = 387$. 
 \end{itemize}
 The second outcome measures how much more prone treated users are to spend money compared to the control. The first is similar but takes into account the spending of each user before the treatment as a means of normalization.

 % Their average uplifts are
%(i) the average money spent by the users after the ``coupon redeem time'';  and (ii) the change of the average money spent before and after it. The former tests how much more prone treated users are to spend money compared to the control, while the latter takes also into account the spending of the user before the treatment as a means of normalization. % Their average uplifts are  
%$1.95$ and $2.6$ respectively.

%\textcolor{blue}{Note that we could not utilize the popular uplift dataset from \textit{Criteo}~\cite{diemert2018large} due to the lack of an inherent underlying graph. While we could create a graph based on feature similarity, it has been proven erroneous for node-level tasks such as the current one~\cite{errica2024class}.}
 %Apart from the standard features, the graph models include a label propagation of the train set outcomes through the graph to assist in semi-supervision.

%The summary statistics of the dataset can be found in Tab. (\ref{tab:tab_stats}). 

%\textcolor{blue}{We have added further experiments with a binary outcome in \ref{sec:appendix}, comparing also with class transformation approaches \cite{rudas2018linear,rudas2023regularization} which are not used in continuous outcomes. }

\subsubsection{MovieLens.} We utilize the \href{https://grouplens.org/datasets/movielens/25m/}{\textsl{Movielens25}} and filter the users based on the minimum number of ratings.% is comprised of 32,848 users, 58,429 movies and 16,063,558 ratings. 
  We extract the features of the movie nodes using a Universal Sentence Encoder-Lite~\cite{cer2018universal} on the concatenation of the title, year, and genres and bring them down to 16 dimensions using principal component analysis. The adjacency is defined as in Eq.~\ref{eq:adj} from the movie-viewer bipartite network.  We define the treatment as $T=1$ if the movie's average rating is over the median of $3.15$ or $T=0$ otherwise, akin to~\cite{ma2022learning,lin2023estimating}. Similar to the same literature, the regression outcome is simulated 5 times: $y_s = \text{ReLU}(w_s x+w_t t + e_s)$, where $w_s \in \mathbb{U}(10,20)$ and $e_s \in \sim \mathcal{N}(10, 5^2)$ are random variables of the simulation.

\begin{table}[t]
    \centering
    \caption{Statistics of the bipartite datasets (directed).}
    \label{tab:tab_stats}
    \begin{tabular}{cccccc}
    \toprule
        Dataset~~~ & ~~~Nodes~~~ &  ~~~$E$ (before $T$)~~~ & $E$ (after $T$) & ~~~~$T=0$~~~~ & $T=1$\\
        \midrule
        \textsl{RetailHero~~} & (180,653+40,542) & 2,522,096 & 12,021,243 & 90,097 &90,556 \\%64,122 & 116,531
        \textsl{MovieLens~~} & (59,047+162,541) & 9,369,966 & 15,630,129 & 29,518 & 29,529 \\%64,122 & 116,531
    \bottomrule
    \end{tabular}
\end{table}

\subsection{Benchmark Models}
To facilitate comparison with the state-of-the-art, we utilize a variety of methodologies from meta learners S, T, X, R~\cite{kunzel2019metalearners} and doubly robust DR~\cite{kennedy2023towards} to uplift trees \textsc{UpliftTree}~\cite{radcliffe2011real}, and neural models \textsc{CEVAE}~\cite{louizos2017causal}, \textsc{Dragonnet}~\cite{shi2019adapting}, and \textsc{CFR}~\cite{shalit2017estimating}. 
We rely on the implementations from the CausalML package~\cite{chen2020causalml} for all, using \textsc{XGBoost} as the output and the build-in \textit{elastic net} as the propensity model whenever required. We include \textsc{NetDeconf}~\cite{guo2020learning} as the graph-aware benchmark from ITE literature but remove the Wasserstein distance cost because it fails to run on a GPU with 32GB. All methods are run with their suggested parameters.
%\url{https://github.com/rguo12/network-deconfounder-wsdm20/tree/master}. }Maybe also these HINITE \cite{lin2023estimating} \url{https://github.com/LINXF208/HINITE}  NetEst\cite{jiang2022estimating} and \cite{veitch2019using}.}
%Furthermore, we develop experiments using the Causalml package for the double ML \cite{kennedy2023towards}, using XGBoost for effect estimation and logistic regression for outcome prediction. 

\subsection{Experiments}

In contrast to the aforementioned studies in ITE~\cite{guo2020learning,ma2021causal}, we can not measure the counterfactual error in the sample level because it does not exist. 
Moreover, since our task is not binary, we can not utilize the uplift curve~\cite{diemert2018large}. Even if we could, our application focuses on the potential of the top ranked users in order to target them with the campaign, hence it is not sensible to focus on the estimation of the whole dataset.
Instead, we rely on the realistic evaluation of our scenario that was also the success criterion for our main dataset\footnote{\url{https://ods.ai/competitions/x5-retailhero-uplift-modeling}.}: %ranking users based on the prediction and computing the real ATE of the set\cite{rafla2022evaluation}.\\ 
\\
\textbf{up@40/20}: We take the top 40\% and 20\% of the test samples sorted based on their predicted uplifts, and measure the real ATE in this set \cite{rafla2022evaluation}.

%\textbf{Up@40/20}: 

\subsubsection{Settings.} %We take the top 40\% and 20\% of the test samples sorted based on their predicted uplifts, and measure the real ATE in this set (\textbf{up@40/20}). %If the ranking is successful, the top samples will have a significant uplift, while the rest will drop. 
To evaluate the models' capacity in semi-supervised generalization, we utilize 5 and 20-fold cross-validation where the test part is set for training and vice versa, i.e., we will have 20\% and 5\% training set size and 5 and 20 iterations, respectively.
We run each method 5 times with random seeds from 0 to 4 and log the average and standard deviation of the respective metric in k-fold validation. For all datasets, we set the learning rate to 0.01 with a weight decay of $1e-4$. We used ReLU as an activation function. The dropout rate is set to 0.4, the number of epochs to 2000, and the hidden layers to (64, 64, 32). Finally, the number of clusters is set to 50 and the coefficients for the acquisition function are 0.2, 0.1, and 0.7, respectively. %The random seeds in each of the 5 runs are 0,1,2,3,4 respectively, .
%The code for the analysis can be found in an anonymous GitHub in Appendix C. 

In the active learning setting, denoted as \textsc{UMGNet-AL}, which utilizes the \textsc{UMGNet-SAGE} model with the proposed acquisition function, instead of 20 and 5-fold, we train an initial model in 1\% and 4\% of the dataset and increase the train set up to 5\%/20\%  respectively with 5 batch queries using a greedy policy.
The experiments are performed with an Nvidia V100 16 GB and 32 GB RAM. The results for \textsl{RetailHero} can be seen in Tables~\ref{tab:results_1}, \ref{tab:results_2}. 
The best result is indicated in \textbf{boldface}, and the second-best is \underline{underlined}.
%\begin{itemize}
    %\item \textcolor{red}{\textbf{MSE}: Since we have ground truth estimations of the uplift outcome, we can compute the accuracy as the squared difference in means $ ( (E[\hat{Y}_t]-E[\hat{Y}_c])-(E[Y_t]-E[Y_c]))^2 $.} %$ ( (\frac{1}{N} (\sum_{i}^N\hat{Y_{i,t} }-\sum_{i}^N\hat{Y{i,c}}) ) - (\frac{1}{N} (\sum_{i}^N Y_{i,t}- \sum_{i}^N Y_{i,c}) ) )^2$.
    %\textcolor{red}{add equation to explain continuous uplift}.
%\end{itemize}

%Note that, the chosen samples have a predefined treatment i.e. we can not change $T$. Moreover, the labeled samples reveal \textbf{new edges} in the graph corresponding to their purchases after they participate in the experiment.\textcolor{red}{Only plots for random policies. Maybe we don't have to add the benchmarks}

\subsubsection{Results and discussion.} 
It can be seen that the proposed methods outperform the benchmarks for both outcome variables. Some benchmarks even exhibit negative uplifts, which means the predicted sets have higher $Y_c$ than $Y_t$. 
The regular $\text{\textit{ATE}}$ is the $\bar{Y}_t-\bar{Y}_c$ of the whole dataset, and it is what we expect to get by a random balanced subset of users without any ranking.
If we consider it as a baseline, we see that in most cases, the benchmarks tend to be under the $\text{\textit{ATE}}$. This is justified by the reduced supervision, which severely diminishes the benchmarks' generalization, moving their results closer to a random sample. In contrast, the proposed methods achieve consistently and sometimes considerably higher uplift than the $\text{\textit{ATE}}$, signifying their ability to generalize in this setting. We actually see that the difference between the proposed methods and the benchmarks is close to or sometimes larger than the actual $\text{\textit{ATE}}$.

\begin{table}[h!]
    \caption{Uplift metrics of the predicted sets in the \textsl{RHC}. Regular $\text{\textit{ATE}} = 2.60$.}
    \label{tab:results_1}
    \centering
    \begin{tabular}{lcccc}
    \toprule
    \multirow{2}{*}{\textsl{Model}} & \multicolumn{2}{c}{20\% training size} & \multicolumn{2}{c}{5\% training size} \\ 
    \cmidrule{2-3} \cmidrule{4-5}
      &  up@40 &  up@20  & up@40 & up@20 \\% & \textbf{MSE} \\
      \midrule
\textsc{S-XGB} & $-1.63 \pm 1.85$ & $-0.67 \pm 3.6$ & 
$1.13 \pm 1.09$ & $2.43 \pm 1.79$	\\%&  15.14888\\
\textsc{T-XGB} & $0.58 \pm 1.48$ & $3.25 \pm 2.72$ & 
$-0.05 \pm 0.32 $ &	$1.33 \pm 1.51$\\ %& 0.150\\
\textsc{X-XGB}& $-2.55 \pm 1.97 $ & $-2.62 \pm 1.47$ & 
$0.21 \pm 1.02$ & $1.19\pm 1.64$	\\%  & 0.000047\\
\textsc{R-XGB}& $1.77\pm 0.97$ & $3.70 \pm 1.62$ & 
$2.01 \pm 0.53$ & $5.2 \pm 1.81$\\
\textsc{DR-XGB} & $-0.26 \pm 1.04$ & $1.18 \pm 1.60$ &
$0.61 \pm 1.00$ & $2.65 \pm 1.05$	\\%  &   24.28294    \\
\textsc{UpliftTree}& $\underline{5.95 \pm 2.43}$ &$2.38 \pm 7.00$ &
$\underline{3.74 \pm 0.77}$ &$2.73 \pm 1.83$\\
%\underline{0.062} & 0.009 &0.0147 & 0.000030\\
\textsc{CFR}& $0.76 \pm 1.36 $ & $-2.09 \pm 1.6$ & $1.73 \pm 0.8$&  $1.41 \pm 1.5$ \\
\textsc{CEVAE}& $3.10 \pm 0.59$ & $3.23 \pm 1.63$  & $2.88\pm 0.75$ & $3.34\pm 0.70$\\
 \textsc{Dragonnet}& $2.24 \pm 0.14$ & $3.08 \pm 0.34$ & $2.26 \pm 0.04$ & $3.01 \pm 0.10$\\% 
\textsc{NetDeconf}&  $1.89 \pm 1.86$ & $2.73 \pm 1.39 $  & $1.08 \pm 0.46 $  & $2.06 \pm 0.62$\\
%GNN &0.161 &-0.099331 \\
%HeteroGNN & - & - &-\\
%HeteroGAT & - & - &-\\
\midrule
%\textsc{UMGNet-NGCF}& $ \underline{5.34 \pm 0.86} $ & $\underline{5.49 \pm 1.15}$ & $3.71 \pm 0.61$&  $ 3.74 \pm 0.97$ \\
%\textsc{UMGNet-LGC}& $4.17 \pm 1.55$ & $3.96 \pm 2.70$ & $3.70 \pm 0.73$&  $ 3.72 \pm 0.87$ \\
\textsc{UMGNet-SAGE}& $3.20 \pm 0.25 $ & $\mathbf{6.48 \pm 0.70}$ & $2.66 \pm 0.75$&  $\underline{5.69 \pm 1.01}$ \\
%\textbf{T-SAGE}$Y_1$& $4.67 \pm 0.41 $ & $\mathbf{8.04 \pm 1.25}$ & $\mathbf{5	\pm 0.06}$& $\mathbf{7.77 \pm 1.01}$  \\
%\textsc{UMGNet-EG} & $6.01\pm 4.33$ & $3.94 \pm 3.18$ & $4.12 \pm 5.00$ & $4.18 \pm 3.00$\\ %
\textsc{UMGNet-AL} & $\mathbf{6.27\pm 3.00}$ & $\underline{4.64 \pm 3.60}$ & $\mathbf{5.83\pm 2.75}$&  $\mathbf{6.83 \pm 3.77}$\\
%\textbf{HeteroSAGE} & & &  & &\\
%\textbf{MLP+Graph} &  & & & &\\
%\textbf{TB-SAGE full fts} & 3.512 & 5.143 &  2.784 & 5.965\\%& 109\\
%\hline
%\textbf{Optimum} & & & &\\
%MultiSAGE &- & -& \\
%SAGE 5 epochs 0.2901630.000028
%XGB+LP & 0.294 & 0 & 0.052
\bottomrule
    \end{tabular}
\end{table}

To be more specific, \textsc{UMGNet-SAGE} is overall more effective in settings with 20\% training size, and \textsc{UMGNet-AL} produces the strongest average uplift when the training size is limited to 5\%, which is sensible due to active learning choosing the most informative training set. 
\textsc{UpliftTree} has the second best performance in \textsl{RHC} albeit with the highest standard deviation ($7$). Furthermore, an effective model should exhibit higher \textbf{up@20} than \textbf{up@40}, meaning the top 20\% predictions should have higher real uplift than the top 40\% if the predictive ranking is consistent. Our methods clearly follow this pattern in 15 out of 16 comparisons, in contrast to \textsc{UpliftTree} in \textsl{RHC}. 
The propensity-based methods, i.e., \textsc{DR-XGB}, \textsc{DragonNet}, \textsc{R-XGB}, are not as accurate because the treatment assignment in the dataset is balanced. Thus, the potential bias is minimized, and accordingly, the effect of the propensity score is diminished.

Throughout the experiments, the standard deviation grows with the average value, with exceptions such as \textsc{UMGNet-SAGE} in \textsl{RHC} and \textsc{T-XGB} in \textsl{RHP}. 
This makes sense given the range of values and their std. 
It should be noted that although both tasks are challenging, \textsl{RHC} has significantly greater std ($521$ to $387$) with a lower average ($266$ to $423$), and it is arguably more informative. \textsl{RHC} normalizes the effect of the treatment with the user's normal behavior, while \textsl{RHP} ranks based on absolute purchase average, which can be biased on the user's preferences. %where we see high average values but lost s.d. %\textsl{RHP} particularly has larger variance between%\textsc{UMGNet-G} is the method with the 

 %This underlines the impact of \textsc{UMGNet}'s uplift gain compared to the rest of the models.}

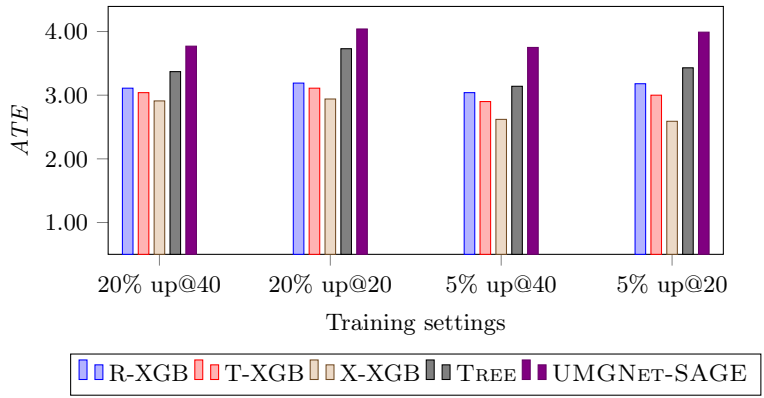
\begin{figure}[t!]
\centering
\begin{tikzpicture}
\begin{axis}[
    ybar,
    bar width=4pt,
    width=0.8\textwidth,
    height=0.4\textwidth,
    yticklabel style={
        /pgf/number format/.cd,
        fixed, fixed zerofill,
        precision=2
    },
    tick align=outside,
    tick pos=left,
    xlabel={Training settings},
    ylabel={\text{\textit{ATE}}},
    ymin=0.5,
    xticklabels={20\% up@40, 20\% up@20, 5\% up@40, 5\% up@20},
    xtick={1, 2, 3, 4},
    legend style={
    at={(0.5,-0.4)},
    legend columns=5,
    anchor=north},
    ]
\addplot coordinates {(1,3.11) (2,3.19) (3,3.04) (4,3.18)};
\addplot coordinates {(1,3.04) (2,3.11) (3,2.9) (4,3)};
\addplot coordinates {(1,2.91) (2,2.94) (3,2.62) (4,2.59)};
\addplot coordinates {(1,3.37) (2,3.73) (3,3.14) (4,3.43)};
\addplot coordinates {(1,3.77) (2,4.04) (3,3.75) (4,3.99)};
\legend{R-XGB,T-XGB,X-XGB,\textsc{Tree},\textsc{UMGNet-SAGE}}
\end{axis}
\end{tikzpicture}
\caption{Uplift of the predicted sets on the \textsl{MovieLens} dataset. Regular $\text{\textit{ATE}} = 0.457$.}
\label{fig:bar_plots}
\end{figure}

Finally, the five realizations of the semi-simulated \textsl{MovieLens} dataset are used to compare \textsc{UMGNet-SAGE} with the best benchmarks from the first experiment. The average results are shown in Fig. \ref{fig:bar_plots}, where it is visible that it consistently outperforms them. In this case, however, the benchmarks perform better compared to \textsl{RetailHero} if we consider the $ATE$, possibly due to the outcome values being smaller and the input features of the movies being considerably more informative.
%Meaning that, the difference between \textsc{UMGNet-SAGE}
%the ATE is important e.g. $1.95$ is the whole ATE so 7.28 - 3.26 is big
%\textit{AC} ATE  $2.60$, average $266$ and sd $521$
%\textit{AP} ATE $1.95$, average $423$ and sd $387$
%clearly outperform the neural models

\begin{table}[t!]
\caption{Uplift metrics of the predicted sets in the \textsl{RHP}. Regular $\text{\textit{ATE}} = 1.95$.}
    \label{tab:results_2}
    \centering
    \begin{tabular}{lcccc}
    \toprule
       \multirow{2}{*}{\textsl{Model}} & \multicolumn{2}{c}{20\% training size} & \multicolumn{2}{c}{5\% training size} \\ 
    \cmidrule{2-3} \cmidrule{4-5}
        & up@40 & up@20  & up@40 & up@20 \\
      \midrule
\textsc{S-XGB} & $3.15 \pm 2.07$  & $4.54 \pm 1.10 $  & 
$2.48 \pm 0.93$	 & $3.56 \pm 0.96$ \\%    16.31664 \\
\textsc{T-XGB} & $ 2.67 \pm 0.26 $ & $\underline{ 5.65 \pm 1.16 }$    &  $2.53 \pm	0.39$ & $4.65 \pm	0.65$ \\ %&  15.45044  \\
\textsc{X-XGB}& $2.96 \pm 1.01 $ & $ 5.33 \pm 1.57 $  &$2.29 \pm	0.35$ & $4.04	\pm 1.06$\\ % &    16.07716\\
\textsc{R-XGB}& $2.66 \pm 1.10$ & $4.58 \pm 0.88$ & $2.90 \pm 0.32$	& $4.11 \pm 0.67$ \\
\textsc{DR-XGB} & $ 3.23 \pm 1.22$  & $3.86 \pm 1.38 $   &$2.77 \pm 0.19$ & $3.61 \pm 1.00$\\%  &   23.9762      \\
\textsc{UpliftTree} & $ 2.84 \pm 0.75$ &$ 5.01\pm 0.45 $ &$1.77 \pm 0.70$ & $3.26 \pm	0.86$ \\
\textsc{CFR} & $1 \pm 1.31$ & $0.25 \pm 1.69$ &  $1.19 \pm 0.55$  &$1.39 \pm 0.81$\\
\textsc{CEVAE}& $2.37\pm 1.67$ & $ 3.26\pm 1.89$ & $ 1.89 \pm 0.54$ & $2.10 \pm 0.54$\\%& \underline{0.062} & 0.009 &0.0147 & 0.000030\
 \textsc{Dragonnet}&  $0.21 \pm 0.16$ & $-0.01 \pm 0.26$  &
$0.21 \pm 0.01$ &$-0.104 \pm 0.01$ \\% 
\textsc{NetDeconf}&$0.84 \pm 1.28$  & $2.00 \pm 1.00$  & $0.45\pm 0.63$ & $0.770 \pm 1.20$\\%& 
%0.257326	4.291299	1.586671	3.772729
\midrule
\textsc{UMGNet-SAGE}& $\mathbf{5.01 \pm 2.16}$ & $\mathbf{7.28 \pm 4.34}$ & $\underline{3.99 \pm 2.00}$&  $\underline{5.07 \pm 3.61}$ \\
%\textsc{UMGNet-EG}& $1.91 \pm2.40$ & $3.59 \pm4.51$ & $3.91 \pm 3.2$ &	$5.66 \pm	3.45$ \\ 
\textsc{UMGNet-AL}& $\underline{ 4.11\pm2.59}$ & $4.69 \pm2.88$ &
$\mathbf{5.89 \pm 2.48}$ &	$\mathbf{6.04 \pm 2.46}$ \\
%\hline
%\textbf{Best w graph} & $3.7 \pm 1.04$	& $7.39 \pm 6.52$ & $2.81 \pm 0.15$	& $5.04 \pm 0.04$\\
%\textbf{TB-SAGE full fts }& $4.08 \pm 1.07$ & $7.6 \pm 3.46$ & 2.7331&    4.62555 \\
%\hline
%\textbf{Optimum} & & & & \\
%MultiSAGE &- & -& \\
%SAGE 5 epochs 0.2901630.000028
%XGB+LP & 0.294 & 0 & 0.052
\bottomrule
    \end{tabular}
\end{table}

% 5_1 4.194354	1.706722	7.649085	4.173442	79.352244	66.116663
% 20_1 4.373761	1.001398	5.010937	2.311416	31.609645	59.967803

% \begin{table}[h!]
% \caption{Summary results on the \textsl{MovieLens} with mean and std. The average treatment effect is $0.457$.}
%     \label{tab:results_3}
%     \centering
%     \begin{tabular}{ccccc}
%     \toprule
%        Model & ~~20\% up@40~~ & ~~20\% up@20~~  & ~~5\% up@40~~ & ~~ 5\% up@20~~ \\
%       \midrule
% %\textbf{S-XGB} &$-0.82\pm 1.07$ & $-0.96\pm1.27$ & $-0.5\pm 1.06$& $-0.5 \pm 1.13$\\%& $3.15 \pm 2.07$  & $4.54 \pm 1.10 $  & &  \\%    16.31664 \\
% \textsc{R-XGB} & $3.11$& $3.19$ & $3.04$ &$3.18$ \\ %&  15.45044  \\
% \textsc{T-XGB}& $3.04$ &$3.11$& $2.9$ & $3$ \\ % &    16.07716\\
% \textsc{X-XGB}&$2.91$	& $2.94$ &$2.62$ & $2.59$\\
% \textsc{Tree}&$3.37$&$3.73$ & $3.14$& $3.43$ \\
% %\textbf{DR-XGB} &$-0.52 \pm 1.09$&  $-0.47 \pm 1.18$& $-0.25 \pm 1.13$& $-0.2\pm 1.21$\\%  &   23.9762      \\
% %\textbf{Dragonnet}&$-0.71\pm0.74$ &$-0.58\pm0.95$ &$-0.55\pm 0.82$& $-0.3\pm 1.05$\\%& 
% \midrule
% \textsc{UMGNet-SAGE}& $3.77$ &$4.04$ & $3.75$ & $3.99$ \\
% %v2 =  23.526667  2.44, 24.035041 2.39
% %\textbf{T-SAGE AL}& & & & \\
% \bottomrule
%     \end{tabular}
% \end{table}

\subsubsection{Ablation study.} We perform a number of extra experiments to answer certain questions of interest.

\paragraph{$\bullet$ Does the GNN help?}
\textsc{Dragonnet} \cite{shi2019adapting} is one of the most popular neural architectures for $\text{\textit{ITE}}$, though it has not been extensively utilized for uplift modeling or in semi-supervised settings. Although they are trained with different parameters, e.g., our architecture has fewer layers, \textsc{UMGNet-Dr} resembles a version of \textsc{Dragonnet} with bipartite SAGE encoding. In Table \ref{tab:results_dragon}, we see that adding information from the network produces significantly better results. The lack of supervision is detrimental for a deep neural network like \textsc{Dragonnet}, as is prevalent in \textsc{RHP}.

\begin{table}[t]
    \caption{The effect of GNN compared to vanilla \textsc{Dragonnet}.}
    \label{tab:results_dragon}
    \centering
    \begin{tabular}{clcccc}
    \toprule
    \multirow{2}{*}{\textsl{Dataset}} & ~~\multirow{2}{*}{\textsl{Model}}& \multicolumn{2}{c}{20\% training size} & \multicolumn{2}{c}{5\% training size} \\ 
    \cmidrule{3-4} \cmidrule{5-6}
     &  & up@40 & up@20  & up@40 & up@20 \\% & \textbf{MSE} \\
      \midrule
\multirow{2}{*}{\textsl{RHC}} & \textsc{Dragonnet}&~ $2.24 \pm 0.14$ & $3.08 \pm 0.34$ & 
$2.26 \pm 0.04$ & $3.01 \pm 0.10$\\% &      nan \\%0.0492 & 
      & \textsc{UMGNet-Dr}&~ $2.41 \pm 1.27$ & $5.20 \pm 1.64$ & $3.00 \pm 0.48$&  $5.70 \pm 0.50$ \\ % AC
      \midrule
      \multicolumn{2}{c}{Improvement (\%)} & +7.6\% & +68.8\% & +32.7\% & +89.4\%
      \\
      \midrule 
      \multirow{2}{*}{\textsl{RHP}} & \textsc{Dragonnet}&~ $0.21 \pm 0.16$ & $-0.01 \pm 0.26$  &~ $0.21 \pm 0.01$ &$-0.10 \pm 0.01$ \\%& 
& \textsc{UMGNet-Dr}&~ $4.19 \pm 1.33$ & $6.15 \pm 2.42$ &~ $3.47 \pm 0.82$&  $4.33 \pm 0.90$ \\ % AP
\midrule
      \multicolumn{2}{c}{Improvement (\%)} & +1895.2\% & +2700\% & +1552.4\% & +4430\%
      \\ 
\bottomrule
    \end{tabular}
\end{table}

\paragraph{$\bullet$ Does the GNN layer impact the prediction?}
We compare the results with GNNs like SAGE~\cite{hamilton2017inductive}, NGCF~\cite{DBLP:conf/sigir/Wang0WFC19}, and LGC~\cite{DBLP:conf/sigir/0001DWLZ020}. These GNN models are used to derive different instances of the \textsc{UMGNet} framework. The interested reader may refer to Appendix B for a detailed presentation of each GNN layer. Note that the layers also differ, i.e., SAGE has only 1 layer, but NGCF has 3.
The results can be seen in Table \ref{tab:results_gnns}, where it is clear that SAGE overall performs better, but NGCF is equally effective in \textsl{RHC}.% and LGC for \textsl{RHP}.

\begin{table}[t!]
    \caption{The effect of different GNN layers in \textsc{UMGNet}.}
    \label{tab:results_gnns}
    \centering
    \begin{tabular}{clcccc}
    \toprule
      \multirow{2}{*}{\textsl{Dataset}}~ & \multirow{2}{*}{\textsc{UMGNet}} & \multicolumn{2}{c}{20\% training size} & \multicolumn{2}{c}{5\% training size} \\ 
    \cmidrule{3-4} \cmidrule{5-6}
      &  & up@40~~ & up@20~~  & up@40~~ & up@20 \\% & \textbf{MSE} \\
      \midrule
\multirow{3}{*}{\textsl{RHC}} & 
\textsc{NGCF}& $ \mathbf{5.34 \pm 0.86} $ & $5.49 \pm 1.15$ & $\mathbf{3.71 \pm 0.61}$&  $ 3.74 \pm 0.97$ \\
& \textsc{LGC}& $4.17 \pm 1.55$ & $3.96 \pm 2.70$ & $3.70 \pm 0.73$&  $ 3.72 \pm 0.87$ \\
& \textsc{SAGE}& $3.20 \pm 0.25 $ & $\mathbf{6.48 \pm 0.70}$ & $2.66 \pm 0.75$&  $\mathbf{5.69 \pm 1.01}$ \\ % AC
      \midrule 
      \multirow{3}{*}{\textsl{RHP}} & 
      \textsc{NGCF} & $ 3.53 \pm 2.30 $ & $ 3.06 \pm 2.51$ & $1.97 \pm 0.24$&  $ 2.22 \pm 0.23$ \\
&\textsc{LGC}& $3.50 \pm 2.00$ & $2.89 \pm 2.25$ & $3.31 \pm 0.52$&  $3.30 \pm 0.77$ \\
&\textsc{SAGE}& $\mathbf{5.01 \pm 2.16}$ &~ $\mathbf{7.28 \pm 4.34}$~~ & $\mathbf{3.99 \pm 2.00}$&  $\mathbf{5.07 \pm 3.61}$ \\
\bottomrule
    \end{tabular}
\end{table}

\paragraph{$\bullet$ Is active learning helpful?}
%Tables \ref{tab:results_1} \& \ref{tab:results_2} facilitate a comparison between the vanilla \textsc{UMGNet-SAGE} and \textsc{UMGNet-AL}, which can be interpreted as the same model under a random policy. 
In Table \ref{tab:results_al} we are utilizing an e-greedy policy, that uses random batches with probability $\epsilon = 0.5$, noted as \textsc{UMGNet-EG}, as a baseline to clarify the need for the acquisition function.
We see a clear improvement in each metric if we optimize the acquisition function in each step. 

\begin{table}[t!]
    \caption{The effect of different active learning policies in \textsc{UMGNet}.}
    \label{tab:results_al}
    \centering
    \begin{tabular}{clcccc}
    \toprule
      \multirow{2}{*}{\textsl{Dataset}}~ & \multirow{2}{*}{\textsc{UMGNet}} & \multicolumn{2}{c}{20\% training size} & \multicolumn{2}{c}{5\% training size} \\ 
    \cmidrule{3-4} \cmidrule{5-6}
   &  & ~~up@40~~ & ~~ up@20~~  & ~~up@40~~ & up@20 \\% & \textbf{MSE} \\
      \midrule
\multirow{2}{*}{\textsl{RHC}} & \textsc{EG}~ & $6.01\pm 4.33$ & $3.94 \pm 3.18$~~ & $4.12 \pm 5.00$ & $4.18 \pm 3.00$\\ %
& \textsc{AL} & $6.27\pm 3.00$ & $4.64 \pm 3.60$ ~~& $5.83\pm 2.75$&  $6.83 \pm 3.77$\\ % AC
\midrule
      \multicolumn{2}{c}{Improvement (\%)} & +4.3\% & +17.8\% & +41.5\% & +63.4\%
      \\ 
      \midrule 
      \multirow{2}{*}{\textsl{RHP}} & \textsc{EG}&~ $1.91 \pm2.40$ & $3.59 \pm4.51$& $3.91 \pm 3.20$ &~	$5.66 \pm	3.45$ \\ 
&\textsc{AL}&~ $4.11\pm2.59$ & $4.69 \pm2.88$ &
$5.89 \pm 2.48$ &~	$6.04 \pm 2.46$ \\\midrule
\multicolumn{2}{c}{Improvement (\%)} & +115.2\% & +30.6\% & +50.6\% & +6.7\%\\
\bottomrule
    \end{tabular}
\end{table}

%% file: conclusion.tex
\section{Conclusion}
\label{sec:conclusion}
The creation of a large enough training set to use for uplift modeling can be a costly, time-consuming, or risky task. It is thus important to develop methodologies that select the right samples to intervene on and extrapolate efficiently on the rest of the dataset. We propose a two-step modular methodology that addresses these needs. 
The main problem is formulated as semi-supervised uplift modeling, and we solve it using bipartite graph neural networks. Additionally, a batch active learning method is defined based on the model's uncertainty, structural importance, and feature diversity to build the training set.

We utilize, for the first time, a large-scale graph with ground-truth experimental information to test our hypothesis. The proposed methodology is compared to a breadth of benchmarks from both uplift modeling and individual treatment effect literature.
Our results indicate a clear advantage of the proposed methodology compared to the benchmarks, which sometimes perform near random in semi-supervised settings. Moreover, active learning enhances the results as the supervision diminishes.
It is important to note that each step can be utilized separately e.g., the acquisition function can be used with other models. This framework aspires to be an initial step towards addressing the realistic yet overlooked problem of uplift modeling under budget for experimental interventions.

Regarding future work, we plan to examine the theoretical aspects of the method. Specifically, we aim to understand better the tradeoff between the number of treated nodes and the generalization capability of the model. Moreover, as the main application revolves around social networks, it is vital to analyze the model's fairness in terms of treatment allocation or outcome prediction.
Finally, we plan to experiment with other available semi-synthetic datasets of varying sizes to research the model's robustness and scalability.

\section{Ethics}

This study only involved public datasets that are freely available for academic purposes. There are no obvious ethical considerations regarding negative impacts from the broader application of the method.

\vspace{.5cm}
\noindent \textbf{Acknowledgements.} Supported in part by ANR (French National Research Agency) under the JCJC project GraphIA (ANR-20-CE23-0009-01).

%\begin{itemize}
%\item Can we use GNNs to identify spillover effects that cause the treatment of a node to interfere with the outcome of another node?
%\item Once you identify spillover effects how to alleviate them from observational studies?
%\item Can we use prediction-powered inference \cite{angelopoulos2023prediction} to update the estimates with confidence intervals?
%\end{itemize}

%% file: appendix.tex
\section*{Appendix}
\appendix
\label{sec:appendix}
\section{Network as a confounder and as a source for interference}
\label{app:interference}
Generally, we can make a distinction between the methods that predict the ATE i) using the network as a confounder~\cite{veitch2019using,guo2020learning,ma2021deconfounding,chu2021graph,jiang2022estimating,farzam2023curvature}, ii) assuming interference bias caused by the network~\cite{ma2021causal,cortez2022staggered,lin2023estimating}, iii) separating between confounding and interference bias~\cite{ma2022learning,cristali2022using}. That said, all methods are predominantly evaluated based on the predicted ITE's accuracy, the main difference being that the models addressing interference predict the causal estimate while setting the message passing to zero. % models do not address how to use the predictions to alleviate the bias from the observed outcome.
Our work falls in the first category and follows the same assumptions as the literature regarding %the absence of 
interference~\cite{veitch2019using,ma2021deconfounding,chu2021graph,farzam2023curvature}. Interference in e-commerce bipartite networks is not studied as extensively as in social networks \cite{ugander2013graph,karrer2021network}. Its existence remains unproved by experimental studies, mainly because it relies on the type of recommendation algorithm and the nature of the network.

\section{GNN layers adopted for the ablation study}
\label{app:gnn}
In the ablation study, we analyzed the possible impact of each GNN layer adopted in our framework, namely, SAGE~\cite{hamilton2017inductive}, NGCF~\cite{DBLP:conf/sigir/Wang0WFC19}, and LGC~\cite{DBLP:conf/sigir/0001DWLZ020}. %We deep dive into the technical details of each of them. 
Hamilton et al.~\cite{hamilton2017inductive} introduce GraphSAGE (abbreviated as SAGE in this work), a GNN able to work on inductive scenarios by predicting labels of nodes unseen in the training set; for each graph node, SAGE's layer samples nodes from the neighborhood and aggregates their features, such as textual representations. Neural Graph Collaborative Filtering~\cite{DBLP:conf/sigir/Wang0WFC19} (NGCF) is one of the first GNNs-based recommendation systems; during the message-passing, the model aggregates the information from the neighbor nodes and calculates the inter-dependencies among the ego and the neighborhood nodes. Light Graph Convolutional network~\cite{DBLP:conf/sigir/0001DWLZ020} (LGC) is a lightweight version of NGCF as it removes node features transformation and non-linearities, inspired by a theoretical observation and empirically validated. %observe that this can lead to superior recommendation accuracy; 
%the model 

\iffalse

\section{Supervised benchmarks}
\label{app:gnn}
\begin{table}[h!]
    \caption{Uplift metrics of the predicted sets in the \textsl{RHC}. Regular $\text{\textit{ATE}} = 2.60$.}
    \label{tab:results_1}
    \centering
    \begin{tabular}{lcccc}
    \toprule
    \multirow{2}{*}{\textsl{Model}} & \multicolumn{2}{c}{50\% training size} & \multicolumn{2}{c}{80\% training size} \\ 
    \cmidrule{2-3} \cmidrule{4-5}
      &  up@40 &  up@20  & up@40 & up@20 \\% & \textbf{MSE} \\
      \midrule
\textsc{S-XGB} &  & & 
 & 	\\%&  15.14888\\
\textsc{T-XGB} &  & & 
&	\\ %& 0.150\\
\textsc{X-XGB}&  & & 
& \\%  & 0.000047\\
\textsc{R-XGB}&  & & 
 & \\
\textsc{DR-XGB} &&  &
& \\%  &   24.28294    \\
\textsc{UpliftTree}&   & &
&\\
 \textsc{Dragonnet}& & &  & \\
\textsc{UMGNet-SAGE}& $3.20 \pm 0.25 $ & $\mathbf{6.48 \pm 0.70}$ & $2.66 \pm 0.75$&  $\underline{5.69 \pm 1.01}$ \\
\bottomrule
    \end{tabular}
\end{table}
\fi 
%\section{Code}
%The code can be found in the anonymous github 
%\url{https://anonymous.4open.science/r/uplift_semisupervised_submission-4570/README.md}.

%% file: main.bbl
\begin{thebibliography}{10}
\providecommand{\url}[1]{\texttt{#1}}
\providecommand{\urlprefix}{URL }
\providecommand{\doi}[1]{https://doi.org/#1}

\bibitem{arbour2016inferring}
Arbour, D., Garant, D., Jensen, D.: Inferring network effects from observational data. In: Proceedings of the 22nd ACM SIGKDD International Conference on Knowledge Discovery and Data Mining. pp. 715--724 (2016)

\bibitem{bakshy2012social}
Bakshy, E., Eckles, D., Yan, R., Rosenn, I.: Social influence in social advertising: evidence from field experiments. In: Proceedings of the 13th ACM Conference on Electronic Commerce. pp. 146--161 (2012)

\bibitem{betlei2021uplift}
Betlei, A., Diemert, E., Amini, M.R.: Uplift modeling with generalization guarantees. In: Proceedings of the 27th ACM SIGKDD Conference on Knowledge Discovery \& Data Mining. pp. 55--65 (2021)

\bibitem{cer2018universal}
Cer, D., Yang, Y., Kong, S.y., Hua, N., Limtiaco, N., John, R.S., Constant, N., Guajardo-Cespedes, M., Yuan, S., Tar, C., et~al.: Universal sentence encoder. arXiv preprint arXiv:1803.11175  (2018)

\bibitem{chen2020causalml}
Chen, H., Harinen, T., Lee, J.Y., Yung, M., Zhao, Z.: Causalml: Python package for causal machine learning. arXiv preprint arXiv:2002.11631  (2020)

\bibitem{chernozhukov2017double}
Chernozhukov, V., Chetverikov, D., Demirer, M., Duflo, E., Hansen, C., Newey, W.: Double/debiased/neyman machine learning of treatment effects. American Economic Review  \textbf{107}(5),  261--265 (2017)

\bibitem{chu2021graph}
Chu, Z., Rathbun, S.L., Li, S.: Graph infomax adversarial learning for treatment effect estimation with networked observational data. In: Proceedings of the 27th ACM SIGKDD Conference on Knowledge Discovery \& Data Mining. pp. 176--184 (2021)

\bibitem{cortez2022staggered}
Cortez, M., Eichhorn, M., Yu, C.: Staggered rollout designs enable causal inference under interference without network knowledge. Advances in Neural Information Processing Systems  \textbf{35},  7437--7449 (2022)

\bibitem{cristali2022using}
Cristali, I., Veitch, V.: Using embeddings for causal estimation of peer influence in social networks. Advances in Neural Information Processing Systems  \textbf{35},  15616--15628 (2022)

\bibitem{dawid1979conditional}
Dawid, A.P.: Conditional independence in statistical theory. Journal of the Royal Statistical Society Series B: Statistical Methodology  \textbf{41}(1),  1--15 (1979)

\bibitem{devriendt2018literature}
Devriendt, F., Moldovan, D., Verbeke, W.: A literature survey and experimental evaluation of the state-of-the-art in uplift modeling: A stepping stone toward the development of prescriptive analytics. Big Data  \textbf{6}(1),  13--41 (2018)

\bibitem{diemert2018large}
Diemert, E., Betlei, A., Renaudin, C., Amini, M.R.: A large scale benchmark for uplift modeling. In: Proceedings of the KDD Workshop on Artificial Intelligence for Computational Advertising (2018)

\bibitem{DBLP:conf/www/Fan0LHZTY19}
Fan, W., Ma, Y., Li, Q., He, Y., Zhao, Y.E., Tang, J., Yin, D.: Graph neural networks for social recommendation. In: {Proceeding of the 28th ACM Web Conference}. pp. 417--426. {ACM} (2019)

\bibitem{farzam2023curvature}
Farzam, A., Tannenbaum, A., Sapiro, G.: Curvature and causal inference in network data. In: Causal Representation Learning Workshop at NeurIPS 2023 (2023)

\bibitem{gal2016dropout}
Gal, Y., Ghahramani, Z.: Dropout as a bayesian approximation: Representing model uncertainty in deep learning. In: International Conference on Machine Learning. pp. 1050--1059. PMLR (2016)

\bibitem{garivier2011upper}
Garivier, A., Moulines, E.: On upper-confidence bound policies for switching bandit problems. In: International Conference on Algorithmic Learning Theory. pp. 174--188. Springer (2011)

\bibitem{gilhuber2023diffusal}
Gilhuber, S., Busch, J., Rotthues, D., Frey, C.M., Seidl, T.: Diffusal: Coupling active learning with graph diffusion for label-efficient node classification. In: Joint European Conference on Machine Learning and Knowledge Discovery in Databases. pp. 75--91. Springer (2023)

\bibitem{graff2021accelerating}
Graff, D.E., Shakhnovich, E.I., Coley, C.W.: Accelerating high-throughput virtual screening through molecular pool-based active learning. Chemical Science  \textbf{12}(22),  7866--7881 (2021)

\bibitem{gui2015network}
Gui, H., Xu, Y., Bhasin, A., Han, J.: Network a/b testing: From sampling to estimation. In: Proceedings of the 24th International Conference on World Wide Web. pp. 399--409 (2015)

\bibitem{guo2020learning}
Guo, R., Li, J., Liu, H.: Learning individual causal effects from networked observational data. In: Proceedings of the 13th International Conference on Web Search and Data Mining. pp. 232--240 (2020)

\bibitem{gutierrez2017causal}
Gutierrez, P., G{\'e}rardy, J.Y.: Causal inference and uplift modelling: A review of the literature. In: Proceedings of the 4th International Conference on Predictive Applications and APIs. pp. 1--13. PMLR (2017)

\bibitem{hamilton2017inductive}
Hamilton, W., Ying, Z., Leskovec, J.: Inductive representation learning on large graphs. Advances in Neural Information Processing Systems  \textbf{30} (2017)

\bibitem{hartford2017deep}
Hartford, J., Lewis, G., Leyton-Brown, K., Taddy, M.: Deep iv: A flexible approach for counterfactual prediction. In: International Conference on Machine Learning. pp. 1414--1423. PMLR (2017)

\bibitem{DBLP:conf/sigir/0001DWLZ020}
He, X., Deng, K., Wang, X., Li, Y., Zhang, Y., Wang, M.: Light{GCN}: Simplifying and powering graph convolution network for recommendation. In: Proceedings of the 43rd ACM SIGIR Conference on Research and Development in Information Retrieval. pp. 639--648. {ACM} (2020)

\bibitem{huang2023conformalized_gnn}
Huang, K., Jin, Y., Candes, E., Leskovec, J.: Uncertainty quantification over graph with conformalized graph neural networks. Advances in Neural Information Processing Systems  \textbf{36} (2023)

\bibitem{jiang2022estimating}
Jiang, S., Sun, Y.: Estimating causal effects on networked observational data via representation learning. In: Proceedings of the 31st ACM International Conference on Information \& Knowledge Management. pp. 852--861 (2022)

\bibitem{johansson2016learning}
Johansson, F., Shalit, U., Sontag, D.: Learning representations for counterfactual inference. In: Proceedings of the 33rdh International Conference on Machine Learning. pp. 3020--3029. PMLR (2016)

\bibitem{karrer2021network}
Karrer, B., Shi, L., Bhole, M., Goldman, M., Palmer, T., Gelman, C., Konutgan, M., Sun, F.: Network experimentation at scale. In: Proceedings of the 27th ACM SIGKDD Conference on Knowledge Discovery \& Data Mining. pp. 3106--3116 (2021)

\bibitem{kennedy2023towards}
Kennedy, E.H.: Towards optimal doubly robust estimation of heterogeneous causal effects. Electronic Journal of Statistics  \textbf{17}(2),  3008--3049 (2023)

\bibitem{kipf2016semi}
Kipf, T.N., Welling, M.: Semi-supervised classification with graph convolutional networks. arXiv preprint arXiv:1609.02907  (2016)

\bibitem{kunzel2019metalearners}
K{\"u}nzel, S.R., Sekhon, J.S., Bickel, P.J., Yu, B.: Metalearners for estimating heterogeneous treatment effects using machine learning. Proceedings of the National Academy of Sciences  \textbf{116}(10),  4156--4165 (2019)

\bibitem{lee2010improving}
Lee, B.K., Lessler, J., Stuart, E.A.: Improving propensity score weighting using machine learning. Statistics in Medicine  \textbf{29}(3),  337--346 (2010)

\bibitem{lin2023estimating}
Lin, X., Zhang, G., Lu, X., Bao, H., Takeuchi, K., Kashima, H.: Estimating treatment effects under heterogeneous interference. In: Joint European Conference on Machine Learning and Knowledge Discovery in Databases. pp. 576--592. Springer (2023)

\bibitem{louizos2017causal}
Louizos, C., Shalit, U., Mooij, J.M., Sontag, D., Zemel, R., Welling, M.: Causal effect inference with deep latent-variable models. Advances in Neural Information Processing Systems  \textbf{30} (2017)

\bibitem{ma2021deconfounding}
Ma, J., Guo, R., Chen, C., Zhang, A., Li, J.: Deconfounding with networked observational data in a dynamic environment. In: Proceedings of the 14th ACM International Conference on Web Search and Data Mining. pp. 166--174 (2021)

\bibitem{ma2022learning}
Ma, J., Wan, M., Yang, L., Li, J., Hecht, B., Teevan, J.: Learning causal effects on hypergraphs. In: Proceedings of the 28th ACM SIGKDD Conference on Knowledge Discovery and Data Mining. pp. 1202--1212 (2022)

\bibitem{ma2021causal}
Ma, Y., Tresp, V.: Causal inference under networked interference and intervention policy enhancement. In: International Conference on Artificial Intelligence and Statistics. pp. 3700--3708. PMLR (2021)

\bibitem{olaya2021or}
Olaya, D., Verbeke, W., Van~Belle, J., Guerry, M.A.: To do or not to do: cost-sensitive causal decision-making. European Journal of Operational Research  \textbf{305(2)},  838–852 (2023)

\bibitem{panagopoulos2023maximizing}
Panagopoulos, G., Tziortziotis, N., Vazirgiannis, M., Malliaros, F.: Maximizing influence with graph neural networks. In: Proceedings of the International Conference on Advances in Social Networks Analysis and Mining. pp. 237--244 (2023)

\bibitem{pearl2009causality}
Pearl, J.: Causality. Cambridge university press (2009)

\bibitem{radcliffe2007using}
Radcliffe, N.: Using control groups to target on predicted lift: Building and assessing uplift model. Direct Marketing Analytics Journal pp. 14--21 (2007)

\bibitem{radcliffe2011real}
Radcliffe, N.J., Surry, P.D.: Real-world uplift modelling with significance-based uplift trees. White Paper TR-2011-1, Stochastic Solutions pp. 1--33 (2011)

\bibitem{rafla2022evaluation}
Rafla, M., Voisine, N., Cr{\'e}milleux, B.: Evaluation of uplift models with non-random assignment bias. In: International Symposium on Intelligent Data Analysis. pp. 251--263. Springer (2022)

\bibitem{rubin1974estimating}
Rubin, D.B.: Estimating causal effects of treatments in randomized and nonrandomized studies. Journal of Educational Psychology  \textbf{66}(5), ~688 (1974)

\bibitem{rubin2005causal}
Rubin, D.B.: Causal inference using potential outcomes: Design, modeling, decisions. Journal of the American Statistical Association  \textbf{100}(469),  322--331 (2005)

\bibitem{rudas2023regularization}
Ruda{\'s}, K., Jaroszewicz, S.: Regularization for uplift regression. In: Joint European Conference on Machine Learning and Knowledge Discovery in Databases. pp. 593--608. Springer (2023)

\bibitem{russo2018tutorial}
Russo, D.J., Van~Roy, B., Kazerouni, A., Osband, I., Wen, Z., et~al.: A tutorial on thompson sampling. Foundations and Trends{\textregistered} in Machine Learning  \textbf{11}(1),  1--96 (2018)

\bibitem{rzepakowski2012decision}
Rzepakowski, P., Jaroszewicz, S.: Decision trees for uplift modeling with single and multiple treatments. Knowledge and Information Systems  \textbf{32},  303--327 (2012)

\bibitem{settles2009active}
Settles, B.: Active learning literature survey  (2009)

\bibitem{shalit2017estimating}
Shalit, U., Johansson, F.D., Sontag, D.: Estimating individual treatment effect: generalization bounds and algorithms. In: Proceedings of the 34th International Conference on Machine Learning. pp. 3076--3085. PMLR (2017)

\bibitem{shi2019adapting}
Shi, C., Blei, D., Veitch, V.: Adapting neural networks for the estimation of treatment effects. Advances in Neural Information Processing Systems  \textbf{32} (2019)

\bibitem{soltys2018boosting}
So{\l}tys, M., Jaroszewicz, S.: Boosting algorithms for uplift modeling. arXiv preprint arXiv:1807.07909  (2018)

\bibitem{stadler2021graph}
Stadler, M., Charpentier, B., Geisler, S., Z{\"u}gner, D., G{\"u}nnemann, S.: Graph posterior network: Bayesian predictive uncertainty for node classification. Advances in Neural Information Processing Systems  \textbf{34},  18033--18048 (2021)

\bibitem{tye2004application}
Tye, H.: Application of statistical ‘design of experiments’ methods in drug discovery. Drug discovery today  \textbf{9}(11),  485--491 (2004)

\bibitem{ugander2013graph}
Ugander, J., Karrer, B., Backstrom, L., Kleinberg, J.: Graph cluster randomization: Network exposure to multiple universes. In: Proceedings of the 19th ACM SIGKDD International Conference on Knowledge Discovery and Data Mining. pp. 329--337 (2013)

\bibitem{vanderschueren2024metalearners}
Vanderschueren, T., Verbeke, W., Moraes, F., Proen{\c{c}}a, H.M.: Metalearners for ranking treatment effects. arXiv preprint arXiv:2405.02183  (2024)

\bibitem{veitch2019using}
Veitch, V., Wang, Y., Blei, D.: Using embeddings to correct for unobserved confounding in networks. Advances in Neural Information Processing Systems  \textbf{32} (2019)

\bibitem{verhelst2023uplift}
Verhelst, T., Petit, R., Verbeke, W., Bontempi, G.: Uplift vs. predictive modeling: a theoretical analysis. arXiv preprint arXiv:2309.12036  (2023)

\bibitem{wager2018estimation}
Wager, S., Athey, S.: Estimation and inference of heterogeneous treatment effects using random forests. Journal of the American Statistical Association  \textbf{113}(523),  1228--1242 (2018)

\bibitem{DBLP:conf/sigir/Wang0WFC19}
Wang, X., He, X., Wang, M., Feng, F., Chua, T.: Neural graph collaborative filtering. In: {SIGIR}. pp. 165--174. {ACM} (2019)

\bibitem{wei2015submodularity}
Wei, K., Iyer, R., Bilmes, J.: Submodularity in data subset selection and active learning. In: Proceedings of the 32nd International Conference on Machine Learning. pp. 1954--1963. PMLR (2015)

\bibitem{wright2006comparing}
Wright, D.B.: Comparing groups in a before--after design: When t test and ancova produce different results. British Journal of Educational Psychology  \textbf{76}(3),  663--675 (2006)

\bibitem{wu2019active}
Wu, Y., Xu, Y., Singh, A., Yang, Y., Dubrawski, A.: Active learning for graph neural networks via node feature propagation. arXiv preprint arXiv:1910.07567  (2019)

\end{thebibliography}
